\documentclass[a4paper]{cas-dc}

\usepackage[numbers,sort&compress]{natbib} 
\usepackage{stfloats}
\usepackage{amssymb}
\usepackage{cleveref}
\usepackage[linesnumbered,ruled,vlined]{algorithm2e}

\def\tsc#1{\csdef{#1}{\textsc{\lowercase{#1}}\xspace}}
\tsc{WGM}
\tsc{QE}
\tsc{EP}
\tsc{PMS}
\tsc{BEC}
\tsc{DE}

\begin{document}
\let\WriteBookmarks\relax
\def\floatpagepagefraction{1}
\def\textpagefraction{.001}

\title [mode = title]{Enhancing Video Memorability Prediction with Text-Motion Cross-modal Contrastive Loss and Its Application in Video Summarization}                      

\tnotetext[1]{This document was supported by the state key development program in 14th Five-Year under Grant No. 2021YFF0900701,2021YFF0602103, 2021YFF0602102, 2021QY1702, and in part by Natural Science Foundation of China (No.61801441)}

\author{Zhiyi Zhu}
\ead{zhuzhiyi@cuc.edu.cn}
\affiliation{organization={Department of Information and Communication Engineering},
            addressline={Communication University of China}, 
            city={Beijing},
            postcode={100020}, 
            state={Beijing},
            country={China}}
\credit{Conceptualization, Methodology, Software, Investigation, Formal Analysis, Writing - Original Draft.}
\fnmark[1]

\author{Xiaoyu Wu}
\cormark[1]
\ead{wuxiaoyu@cuc.edu.cn}
\credit{Conceptualization, Funding Acquisition, Resources, Supervision, Writing - Review \& Editing.}
\fnmark[2]

\author{Youwei Lu}
\ead{wowanglenageta@sina.com}
\credit{Data Curation, Writing, Visualization, Supervision,  Writing - Review \& Editing.}
\fnmark[3]

\cortext[cor1]{Corresponding author}

\begin{abstract}
Video memorability refers to the ability of videos to be recalled after viewing, playing a crucial role in creating content that remains memorable. Existing models typically focus on extracting multimodal features to predict video memorability scores but often fail to fully utilize motion cues. The representation of motion features is compromised during the fine-tuning phase of the motion feature extractor due to a lack of labeled data. In this paper, we introduce the Text-Motion Cross-modal Contrastive Loss (TMCCL), a multimodal video memorability prediction model designed to enhance the representation of motion features. We tackle the challenge of improving motion feature representation by leveraging text description similarities across videos to establish positive and negative motion sample sets for a given target. This enhancement allows the model to learn similar feature representations for semantically related motion content, resulting in more accurate memorability predictions. Our model achieves state-of-the-art performance on two video memorability prediction datasets. Moreover, the potential applications of video memorability prediction have been underexplored. To address this gap, we present Memorability Weighted Correction for Video Summarization (MWCVS), using video memorability prediction to reduce subjectivity in video summarization labels. Experimental results on two video summarization datasets demonstrate the effectiveness of MWCVS, showcasing the promising applications of video memorability prediction.
\end{abstract}

\begin{keywords}
video memorability prediction \sep contrastive loss \sep cross-modal \sep video summarization
\end{keywords}

\maketitle

\section{Introduction}

\begin{figure}[ht]
	\centering
	\includegraphics[width=.9\columnwidth]{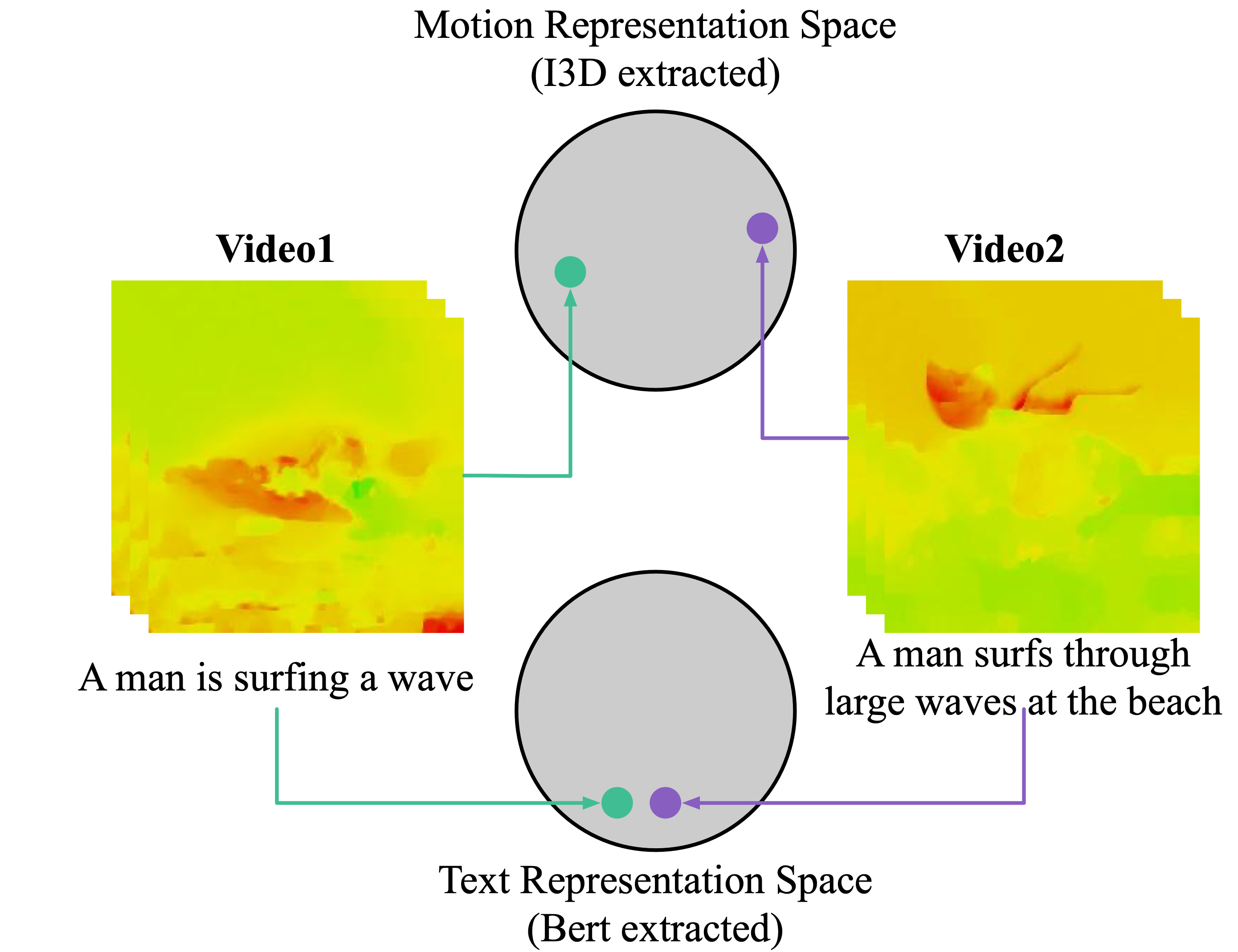}
	\caption{A diagram is presented, illustrating optical flow images from two videos, along with their corresponding text descriptions. Both images depict individuals surfing on a woven surface. The motion and text semantics in both instances align. However, the limited availability of labeled data during the fine-tuning phase of the motion feature extractor results in weak motion features. As a result, the motion features are inconsistent in the motion representation space (as shown in the figure), while the text features (depicted as more consistent in the figure) maintain semantic coherence.}
	\label{FIG:1}
\end{figure}

Humans exhibit diverse memorability levels in response to various stimuli. Prior research has established a correlation between human memory and the perceptual information individuals derive from these stimuli, encompassing both visual and text-based elements \cite{steel2021network}. This association suggests that memorability is intricately linked to content and can be predicted based on its intrinsic characteristics. While memorability has been a longstanding subject of study in psychology, exemplified by the seminal Ebbinghaus curve \cite{ebbinghaus2013memory}, it is a relatively nascent domain in computer vision research. Image memorability served as the pioneering focus in this field, defined as the probability that an image is recalled \cite{isola2011makes}. The literature \cite{fajtl2018amnet,squalli2018deep} has achieved performance approximating human consistency in this task. Experimental results illustrate that machine vision can effectively capture media memorability and its correlation with media content. \cite{isola2011makes} regarded image memorability as an intrinsic property of images. Simultaneously, video has become ubiquitous in our daily lives, and leveraging video memorability can aid in identifying memorable segments, consequently excluding less memorable content. This spans various domains, including video summarization and storytelling \cite{lu2022video}.

Video memorability, defined as the extent to which videos are recalled after a specified duration \cite{cohendet2019videomem}, is commonly evaluated through a video memorability score. This score, expressed numerically on a scale from 0 to 1, signifies greater ease of recall with higher values and increased difficulty in remembering the video with lower values. Previous studies \cite{cohendet2019videomem,2018Annotating,newman2020multimodal} extended the concept of memorability to videos by introducing video memorability prediction datasets and developing models for predicting video memorability scores. Notably, existing video memorability prediction models have predominantly focused on extracting robust visual appearance features \cite{cohendet2019videomem,constantin2022aimultimedialab,li2022adaptive,newman2020multimodal,sweeney2022diffusing,usmani2022modelling}, overlooking the significance of visual motion features. Visual motion cues play a crucial role in media memorability. While \cite{newman2020multimodal} extracted motion features by fine-tuning I3D \cite{carreira2017quo}, the limitation of data insufficiency led to the ordinary performance of the fine-tuned model. In this paper, we introduce the Text-Motion Cross-modal Contrastive Loss (TMCCL) multimodal video memorability prediction model to address the issue of insufficient training data in fine-tuning the motion extractor. TMCCL incorporates text descriptions to establish positive and negative motion sample sets, enhancing the discriminative capability of motion features, as illustrated in Figure 1. This approach results in a robust representation of motion content, thereby improving the accuracy of video memorability score prediction.

Video summarization seeks to condense lengthy videos by selecting engaging and captivating clips. While existing models for video summarization have predominantly focused on developing intricate networks or incorporating diverse categories of features \cite{ghauri2021supervised, lin2022deep, liu2022video, zhao2021reconstructive, zhu2022learning}, they have tended to neglect the influence of human cognition and perception on this process. Defining what constitutes interest and fascination in a video proves challenging, as the subjective nature of judgment is closely tied to human perception and cognition. The level of appeal to human observers directly correlates with the memorability of the content \cite{khosla2015understanding}. Consequently, understanding the intricate interplay between human cognitive processes and video content is essential for advancing video summarization.In this paper, we expand the scope of video memorability prediction to encompass video summarization. We introduce a novel approach termed Memorability Weighted Correction for Video Summarization (MWCVS). Consequently, video clips exhibiting elevated memorability scores are typically more appealing to human observers and are predisposed to constitute a more compelling video summary.

The contributions of this paper are summarized as follows:
\begin{itemize}
    \item We propose a Text-Motion Cross-modal Contrastive Loss (TMCCL) multimodal video memorability prediction model to overcome the challenge of insufficient training data in motion feature extractor fine-tuning, which improves the representation of motion features.

    \item We propose Memorability Weighted Correction for Video Summarization (MWCVS) to solve the problem of human subjectivity in video summarization labels.

    \item Our proposed TMCCL multimodal video memorability prediction model achieves the newly state-of-the-art performance on two related datasets. And our proposed MWCVS also achieves better performance on two video summarization datasets.
\end{itemize}

\section{Related work}

We present the related work in three fields: video memorability prediction, contrastive learning and video summarization. 

\subsection{Video memorability prediction.} 
The paucity of datasets and evaluation metrics constitutes a significant impediment to the advancement of video memorability prediction. Various collection protocols have been devised to construct datasets, such as fMRI \cite{han2014learning} and questionnaires \cite{2018Annotating}. Noteworthy contributions in the literature \cite{cohendet2019videomem,newman2020multimodal} have introduced large-scale datasets and established an objective protocol for quantifying video memorability scores. Among the influencing factors, visual cues emerge as the most pivotal in shaping video memorability. Literature sources \cite{sweeney2020predicting,usmani2022modelling} have leveraged 2-D convolutional neural network (CNN), exemplified by ResNet \cite{he2016deep} and DenseNet \cite{huang2017densely}, to extract visual appearance features. Additionally, 3-D CNN, including ResNet3D \cite{hara2018can} and I3D \cite{carreira2017quo}, have been employed for feature extraction, as demonstrated by the works of \cite{cohendet2019videomem,reboud2021exploring}. More recently, transformer-based representation learning models have achieved significant breakthroughs. The study by \cite{kleinlein2021thau} employed Beit \cite{bao2021beit} to extract visual appearance features, while Deit \cite{touvron2021training} was utilized in the work of \cite{constantin2021using}. Considering videos with accompanying text descriptions, text cues have been employed to enhance model performance through methods such as decision-level fusion \cite{newman2020multimodal} or video caption generation \cite{cohendet2019videomem}. Notably, prior investigations have largely overlooked the impact of motion cues on video memorability, and the inadequate training data in extractor fine-tuning has resulted in suboptimal representation of motion features.

\subsection{Contrastive learning.} 
Recent advancements in self-supervised learning have highlighted the advantages of employing discriminative contrastive loss \cite{caron2021emerging, xinlei2021anempirical, radford2021learning, tao2023siamese}. This approach seeks to generate supervised information for model training in the absence of labels, presenting a viable solution to the issue of insufficient training data during the fine-tuning of models. Specifically, the objective for a target sample is to distinguish its transformed version from other sets of samples \cite{qian2021spatiotemporal, ranasinghe2022self}. The construction of appropriate positive and negative samples is crucial. For instance, SVT \cite{ranasinghe2022self} established global and local spatiotemporal views with multiple spatial scales and frame rates. Moreover, the benefits of constructing cross-modal positive and negative samples were demonstrated \cite{han2020self, morgado2021audio}, where one view learns complementary information from the other. CoCLR \cite{han2020self} utilized motion and visual appearance cues to generate positive and negative samples, leveraging the complementary cues provided by the alternative view. AVID \cite{morgado2021audio} employed contrastive learning for the cross-modal distinction of video and audio. Building on the principles of cross-modal contrastive learning, we introduce TMCCL, which incorporates a text view to provide complementary cues to the motion view. Simultaneously, the motion feature extractor is fine-tuned to enhance the representation of motion features.

\subsection{Video summarization.} 
Effectively accessing noteworthy or pivotal segments within videos not only minimizes human energy expenditure but also accelerates the process of information retrieval. Video summarization, a pivotal task, involves generating a succinct and all-encompassing summary to encapsulate the essence of an extended video. Presently, predominant models in the field center around deep learning techniques, as evidenced by works such as \cite{ghauri2021supervised,lin2022deep,liu2022video,zhao2021reconstructive,zhu2022learning}. Noteworthy contributions in the literature, specifically \cite{lin2022deep}, have established a hierarchical Long Short-Term Memory (LSTM) network tailored to capture temporal dependencies. Additionally, \cite{ghauri2021supervised} employs an attention mechanism, incorporating three distinct sources of features to analyze visual and motion content. However, prevailing methodologies have predominantly concentrated on refining network structures or integrating multi-source features to enhance model performance, often overlooking the crucial aspects of human perception and cognition. Consequently, our proposed method, MWCVS, uniquely introduces considerations for human perception and cognition into the video summarization process.

\section{Methodology}

In this section, we introduce two models: (1) the TMCCL multimodal video memorability prediction model and (2) the Memorability Weighted Correction for Video Summarization (MWCVS). An overview of our proposed video memorability prediction model is shown in Figure 2. Section 3.1 sequentially outlines Text Embeddings, Visual Appearance Embeddings based on Multi-level Encoding, and Text-Visual Appearance Attention, followed by Motion Embeddings based on TMCCL and Decision-Level Fusion. Section 3.2 introduces the video summarization baseline (MSVA) and the enhanced MSVA incorporating MWCVS.

\subsection{TMCCL Multimodal Video Memorability Prediction Model}

\subsubsection{Text Embeddings}

Each video is accompanied by textual descriptions that delineate the semantic content of the video. We employ the pre-trained BERT model \cite{kenton2019bert}, renowned for its remarkable zero-shot transfer learning capability, to extract features at the sentence level. Our processing involves the removal of punctuation, conversion of capital letters to lowercase, and word splitting operations for each sentence. Subsequently, a [CLS] token is appended to the commencement of each sentence, encapsulating the semantic information of the entire sentence. Ultimately, we derive the text features $f_t \in \mathbb{R}^{D_t}$ by extracting the [CLS] token from the last transformer layer, where $D_t$ denotes the dimension of the text feature.

\begin{figure*}
	\centering
	\includegraphics[width=.9\textwidth]{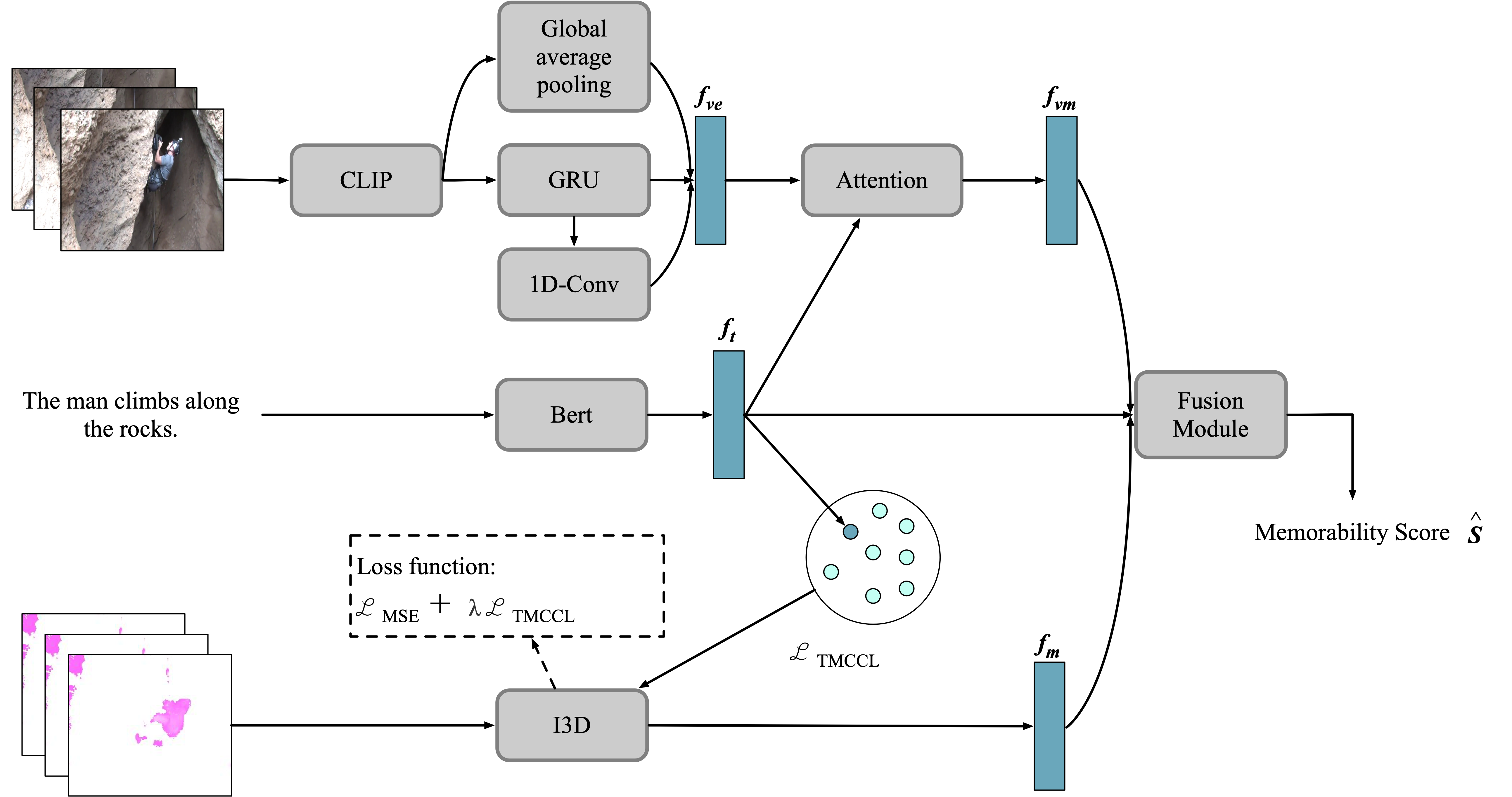}
        \caption{Overview of TMCCL multimodal video memorability prediction model.}
	\label{FIG:2}
\end{figure*}

\subsubsection{Visual Appearance Embeddings based on Multi-level Encoding and Text-visual appearance Attention}

Previous works have leveraged 2D CNN-based or 3D CNN-based networks to extract visual appearance features. Recently, significant advancements have been made with transformer-based large-scale pre-trained models \cite{dosovitskiy2020image,liu2021swin,ranasinghe2022self}, showcasing robust representation learning capabilities. In our approach, we employ CLIP \cite{radford2021learning} to extract frame-level features. To aggregate temporal information, a global average pooling layer is typically incorporated. The work by Dong et al. \cite{dong2021dual} introduced a multi-level encoding method for videos, progressively capturing global, local, and temporal information. We adopt a similar method to encode frame-level features extracted by CLIP, yielding notable advantages. However, we posit that the contribution of distinct feature levels to video memorability prediction is uneven. Consequently, we introduce a cross-modal text-visual appearance attention method to bolster the representation of visual appearance features.

In the context of our study, we adopt a systematic approach to processing input videos. Specifically, a set of $n$ frames is uniformly sampled from the video. Subsequently, each frame is individually fed into the CLIP model, resulting in the extraction of frame features denoted as $\{v_1, v_2, ..., v_n\}$, where $v_i$ represents the deep feature vector of the $i$-th frame. To mitigate frame redundancy, we empirically set $n$ to 8.

Global features $f_{v1}$ are directly derived through global average pooling. This technique facilitates the capture of visual semantics that manifest repeatedly across various frames. Mathematically, this process can be expressed as:

\begin{equation}
    f_{v1} = \frac{1}{n}\sum_{i=1}^nv_i
    \label{equaiton1}
\end{equation}

The frame features are independent and lack interaction. The bidirectional Gated Recurrent Unit (GRU) is renowned for its efficacy in addressing sequence-related challenges by considering both past and future contextual cues. In this context, it is employed to model temporal information within video data. The vectors $\{v_1, v_2, ..., v_n\}$ serve as inputs to a bidirectional GRU, comprising two distinct GRUs identified as $GRU_{forward}$ and $GRU_{backward}$. $GRU_{forward}$ encodes data in the standard order, while $GRU_{backward}$ encodes data in reverse order. These components are denoted as follows:

\begin{equation}
    \overrightarrow{h_i} = GRU_{forward}(\overrightarrow{h_{i-1}}, \overrightarrow{v_i})
    \label{equation2}
\end{equation}

\begin{equation}
    \overleftarrow{h_i} = GRU_{backward}(\overleftarrow{h_{i-1}}, \overleftarrow{v_{n+i-1}})
    \label{equation3}
\end{equation}
where $\overrightarrow{h_i}$ and $\overleftarrow{h_i}$ denote the $i$-th time hidden state in $GRU_{forward}$ and $GRU_{backward}$, respectively. The dimension of hidden state vectors is set to 1024. $\overrightarrow{h_i}$ and $\overleftarrow{h_i}$ are concatenated to obtain bi-GRU output $h_i = [\overrightarrow{h_i}, \overleftarrow{h_i}]$. Gathering all $h_i$ together, we obtain $H = \{h_1, h_2, ..., h_n \}$. Then, we obtain the temporal features with average pooling on $H$. It is represented as:

\begin{equation}
    f_{v2} = \frac{1}{n}\sum_{i=1}^nh_i
    \label{equation4}
\end{equation}

We employ a one-dimensional Convolutional Neural Network (1-D CNN) to augment local details. To capture multi-scale information, we leverage 1-D CNN with varying kernel sizes denoted by $k$. Specifically, we employ kernel sizes of 2, 3, 4, and 5, with a total of 512 kernels in our model. The output vectors from the 1-D CNN undergo padding to ensure uniform dimensions. Subsequently, ReLU activation and mean pooling are applied to compress the vectors. This sequential process is delineated as follows:

\begin{equation}
    m_k = {\rm average\_pooling(ReLU(Conv1d}(H)))
    \label{equation5}
\end{equation}

The vectors $m_k$ are concatenated to obtain the local features $f_{v3}$,
that is:

\begin{equation}
    f_{v3} = [m_2, m_3, m_4, m_5]
    \label{equation6}
\end{equation}

Finally, we concatenate three levels features to obtain multi-level
visual appearance features $f_{vm} \in R^{Dvm}$, that is:

\begin{equation}
    f_{vm} = [f_{v1}, f_{v2}, f_{v3}]
    \label{equation7}
\end{equation}

Next, we employ a cross-modal text-visual appearance attention method to assess the significance of various levels of features within the function $f_{vm}$ concerning the prediction of video memorability. Previous work, as evidenced in \cite{fajtl2018amnet}, has demonstrated the efficacy of attention mechanisms in the context of media memorability. The function $f_{vm}$ is uniformly divided into $l$ segments, thereby establishing:

\begin{equation}
    f_{vm} = [f_{vm}^1, f_{vm}^2, ..., f_{vm}^l]
    \label{equation8}
\end{equation}
where $f_{vm}^i$ is $[(i-1)\times(D_{vm}/l)+1, i\times(D_{vm}/l)]$ dimensions of $f_{vm}$. $l$ is set to 9.

$f_t$ and $f_{vm}^i$ are projected into a common semantic space by several
 linear layers, represented as:

\begin{equation}
    \widehat{f^i}_{vm} = W_v({\rm ReLU}(U_v(f_{vm}^i)))
    \label{equation9}
\end{equation}

\begin{equation}
    \widehat{f_t} = W_t({\rm ReLU}(U_t(f_t)))
    \label{equation10}
\end{equation}
where $U_v$, $W_v$, $U_t$, $W_t$ are weight parameters in linear layers.

$e_i$ is obtained by a element-wise sum of $\widehat{f^i}_{vm}$ and $\widehat{f_t}$, that is:

\begin{equation}
    e_i = W({\rm tanh}(\widehat{f^i}_{vm} + \widehat{f_t}))
    \label{equation11}
\end{equation}
where $W$ is weight parameters of linear layers.

$e_i$ is projected into probability space, where $\sum_{i=1}^l\alpha_i$ = = 1. $\alpha_i$ indicates the importance of $f_vm^{i}$ to video memorability, which is produced by a softmax layer.

\begin{equation}
    \alpha_i = \frac{{\rm exp}(e_i)}{\sum_{j=1}^l{\rm exp}(e_j)}
    \label{equation12]}
\end{equation}

The enhanced visual appearance features $f_{ve}$ is produced by weighted sum as follows:

\begin{equation}
    f_{ve} = \sum_{i=1}^l \alpha_i f_{vm}^i
    \label{equation13}
\end{equation}

\subsubsection{Motion Embeddings based on TMCCL}

\begin{figure}[ht]
    \centering
    \includegraphics[width=\linewidth]{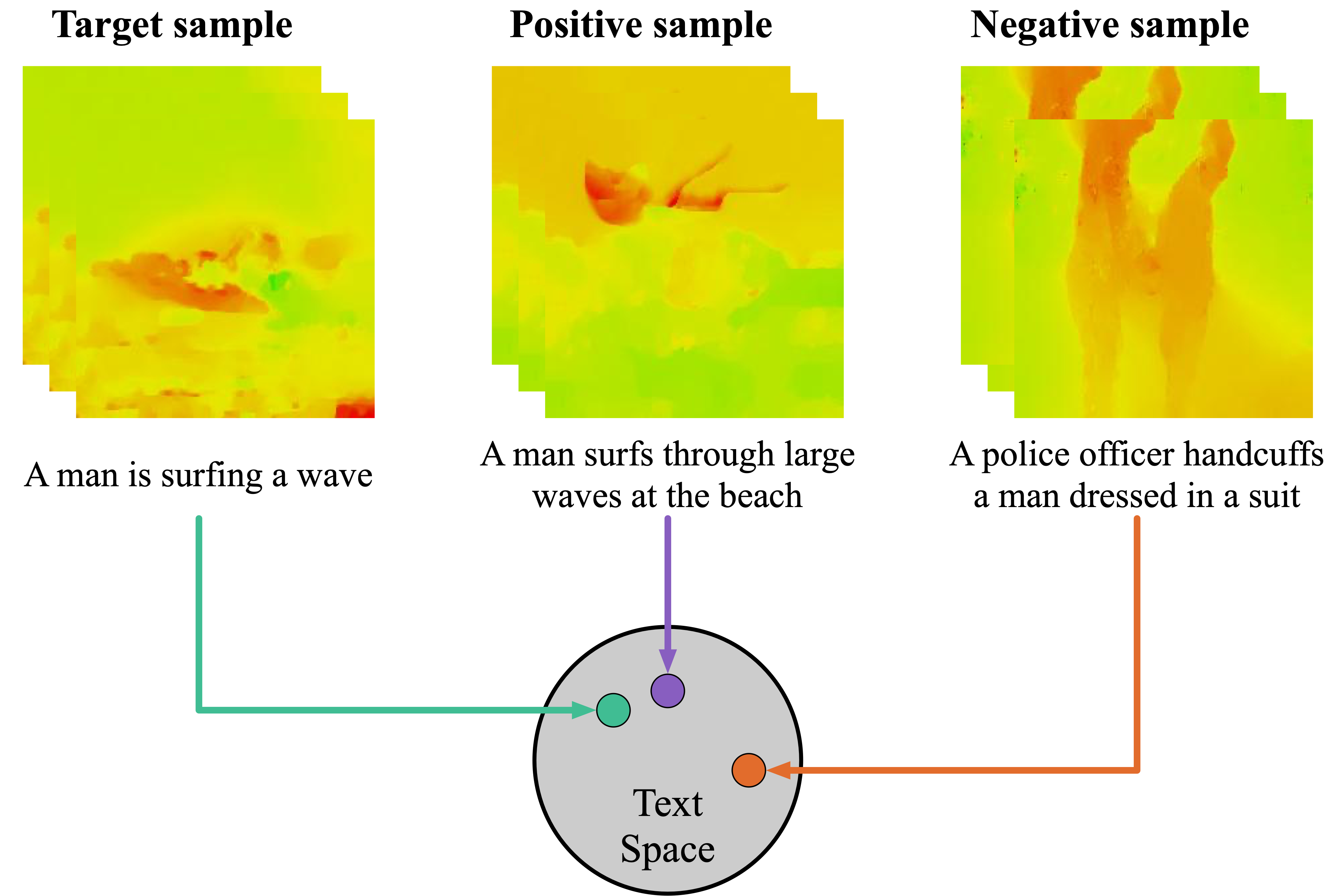}
    \caption{The example of constructing positive and negative samples for a target sample based on text space similarity. The target sample has a high similarity to the positive sample in text space, while the target sample has a low similarity to the negative sample in text space.}
    \label{figure3}
\end{figure}

We employ the pre-trained I3D to extract motion features, denoted as $f_m$, for the representation of visual motion information. The I3D model is exclusively pre-trained on the Kinetics dataset (240k), possessing a substantial number of parameters in its 3D-CNN, thereby incorporating a pronounced inductive bias. In contrast, CLIP and BERT undergo pretraining on datasets comprising 400 million and 3300 million instances, respectively. These extensive datasets contribute to their robust zero-shot transfer capabilities, enhancing the resilience and effectiveness of the features they generate.

Fine-tuning of models is a widely employed paradigm for enhancing feature representation. In our paper, we opt to fine-tune the I3D model using video memorability datasets. However, the inadequacy of data for model fine-tuning results in a suboptimal representation of motion features in the context of video memorability prediction. To address this limitation, we aim to augment the information available for model fine-tuning through the incorporation of text cues to augment motion features. This augmentation is realized by applying Text-Motion Cue Contrastive Learning (TMCCL) to the fine-tuning process of I3D, wherein positive and negative samples are constructed based on text cues.

The introduction of TMCCL involves considering two views, denoted as {$f_m$, $f_t$}, for a given target sample, where $f_m$ and $f_t$ represent motion features (I3D) and text features (Section 3.1.1), respectively. We define $P_im$ and $N_im$ as the sets of positive and negative samples for the target sample. The construction of $P_im$ and $N_im$ is elucidated in Figure 3. Specifically, we compute the similarity between the text features of the target sample and other samples, selecting several samples with the highest similarity (close in Figure 3) as positive samples, and randomly choosing several untaken samples (distant in Figure 3) as negative samples. The detailed procedural steps are outlined in Algorithm 1. Ultimately, we define TMCCL as follows:

\begin{equation}
    \mathcal L_{TMCCL}= {\rm log}\frac{\sum_{f_m^+ \in P_{im}}{\rm sim}(f_m,f_m^+)}{\sum_{f_m^+ \in P_{im}}{\rm sim}(f_m,f_m^+) + \sum_{f_m^- \in N_{im}}{\rm sim}(f_m,f_m^-)}
    \label{equation14}
\end{equation}
where ${\rm sim}(f_m,f_m^+)$ means the similarity calculation between $f_m$
and $f_m^+$, denoted as:

\begin{equation}
    {\rm sim}(f_m,f_m^+) = {\rm exp}(f_m,f_m^+/\tau)
\end{equation}

For model architecture, we introduce a projection head designed for similarity calculation and a regression head tasked with predicting the video memorability score subsequent to the Mixed\_5c layer of the I3D model. The projection head comprises a 3D-CNN, ReLU, and another 3D-CNNs. This configuration, a widely adopted practice in contrastive learning, serves the purpose of mitigating the risk of losing crucial information during feature similarity computation, thereby preserving more comprehensive information in the features before projection. On the other hand, the regression head is composed of a linear layer, Dropout, another linear layer, and a Sigmoid activation function. The composite loss function employed for fine-tuning the I3D model is delineated as follows:

\begin{equation}
    \label{equation16}
    \mathcal L_{MSE} = (\hat{S}_m-S)^2
\end{equation}

\begin{equation}
    \label{equation17}
    \mathcal L_{overall} = \mathcal L_{MSE} + \lambda \cdot \mathcal L_{TMCCL}
\end{equation}
where $\hat{S}_m$ is the score predicted by I3D, $S$ is the ground-truth and $\lambda$ controls the degree of TMCCL.

\begin{algorithm}[ht]
    \caption{The process of consturction of positive and negative samples for TMCCL}
    \label{algorithm1}
    \KwIn{Text descriptions of all samples $\mathbb{C}$; Model $M$. 
    
    Extract text features $f_t$ of all samples based on Bert.}

    \ForEach{batch}{
    calculate text similarity scores between target sample and other samples in $\mathbb{C}$;

    retrieve 2$K$ samples with highest scores as the latent set;

    randomly select $K$ samples from latent set as $P_{im}$, defined as:
    
    $P_im = \{f_{m\_i},f_{m\_k} | \in {\rm Ramdom}K({\rm top2}K(f_{t\_i} \cdot f_{t\_j})), \forall j \in \mathbb{C} \}$;

    use dynamic queue $\mathbb{Q}$ (Moco \cite{he2020momentum}) to build negative sample pool and treat the samples in queue as $N_{im}$, defined as:
    
    $N_{im} = \{f_{m\_j} | \forall f_{m\_j} \notin P_{im} \& j \in  \mathbb{Q} \}$;

    train model $M$ based on $P_{im}$ and $N_{im}$ and update queue.
    }
\end{algorithm}

\subsubsection{Decision-Level Fusion}

Decision-level fusion has demonstrated robust performance in the realm of video memorability prediction, as highlighted in studies such as \cite{li2022adaptive,newman2020multimodal}. In our paper, we employ a decision-level fusion strategy based on self-adaptive weights to effectively integrate visual appearance features ($f_{ve}$), text features ($f_t$), and motion features ($f_m$). These feature sets are input into three MLPs, each comprising a linear layer, ReLU, and another linear layer. The purpose of these MLPs is to predict corresponding scores, denoted as $\hat{S}_v$, $\hat{S}_t$, and $\hat{S}_m$.

Let $\theta_v$ , $\theta_t$, $\theta_m$ be the score weights of $\hat{S}_v$, $\hat{S}_v$ and $\hat{S}_v$ and we define the following limits:

\begin{equation}
    \theta_v + \theta_t + \theta_m = 1
    \label{equation18}
\end{equation}

\begin{equation}
    \theta_v = 1 - t_v \cdot c
    \label{equation19}
\end{equation}

\begin{equation}
    \theta_m = 1 - t_m \cdot c
    \label{equation20}
\end{equation}
where $c$ is the step size (0.05 in our experiment) and $t_v$,$t_m$ vary
among $[0, 1, ..., 1/c]$.

The final predicted score $\hat{S}$ is obtained by weighted sum, that is:

\begin{equation}
    \label{equation21}
    \hat{S} = \theta_v \cdot \hat{S}_v + \theta_t \cdot \hat{S}_t + \theta_m \cdot \hat{S}_m
\end{equation}

\begin{figure*}[ht]
    \centering
    \includegraphics[width=\linewidth]{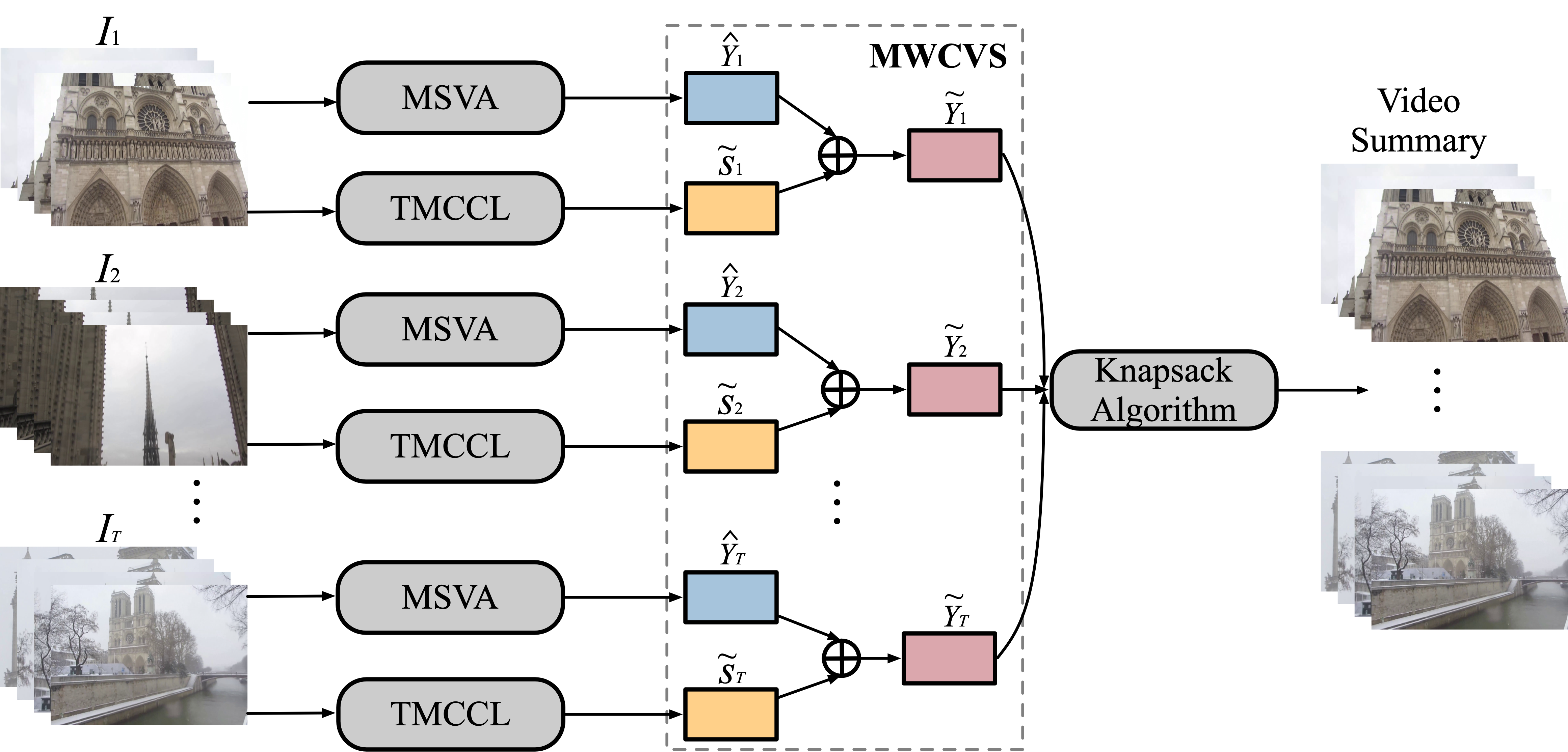}
    \caption{Overview of the Memorability Weighted Correction for Video Summarization model.}
    \label{figure4}
\end{figure*}

\subsection{Memorability Weighted Correction for Video Summarization}

\subsubsection{Multi-Source Visual Attention Model}

A panoramic view of video summarization model with MWCVS is shown in Figure 4. It is noteworthy that our enhancements are applied to the pre-existing MSVA model \cite{ghauri2021supervised}, renowned for its commendable performance in video summarization. Within this section, we commence by elucidating the data flow within MSVA and expounding on the model's output.

The extended video, denoted as $I$, undergoes segmentation into multiple clips employing the split method stipulated by \cite{song2015tvsum}, resulting in a set of clips: $I = \{I_1, I_2, ..., I_T\}$, where $I_i$ represents the $i$-th clip. It is essential to highlight that the chosen split method ensures equitable treatment of the content. Each clip, denoted as $I_i$, is fed into the MSVA model, yielding a corresponding summary importance score $\hat{Y}_i$. This process is repeated for all clips, generating a sequence of summary importance scores denoted as $\hat{Y} = \{\hat{Y}_1, \hat{Y}_2, ..., \hat{Y}_T\}$. While a detailed exposition of the MSVA architecture is omitted in this context, its intricate design and functionality form an integral part of the subsequent sections. 

\begin{equation}
    \label{equation22}
    \hat{Y}_i = MSVA(I_i)
\end{equation}

Finally, we need to select clips that have the highest summary importance scores with a total frame number not exceeding 15\% of the original video to form the summary. This can be considered as a classical 0-1 knapsack problem and be solved by dynamic programming.

\subsubsection{MSVA with MWCVS}

Video summarization presents a subjective challenge owing to the inherent difficulty in precisely delineating the captivating and noteworthy segments within a video. Consideration of human cognitive and perceptual factors, such as video memorability, is justified. Notably, individuals can retain recollections of video clips that engage their interest for an extended duration following a single viewing \cite{konkle2010scene}. Conversely, video clips lacking in appeal are challenging to commit to memory, even after repeated exposure. Cumulatively, these observations underscore video memorability as an indication of human cognitive and perceptual processes. Video clips endowed with higher memorability scores exhibit greater allure to viewers and are predisposed to contribute more effectively to a video summary.

As depicted in Figure 4, each clip, denoted as $I_i$, undergoes input into TMCCL (see Section 3.1), and the corresponding video memorability score, denoted as $\hat{S}_i$, is predicted. We employ a TMCCL model pretrained on the Memento10k dataset (introduced in the next section), with removal of text-related modules from TMCCL. This modification aligns with our objective of leveraging video memorability in application. The rectified summary importance score, denoted as $\Tilde{Y}_t$, is computed through a weighted sum, specifically:

\begin{equation}
    \label{equation23}
    \hat{S}_i = TMCCL(I_i)
\end{equation}

\begin{equation}
    \label{equation24}
    \Tilde{Y}_t = \hat{Y}_i + \mu \cdot \hat{S}_i 
\end{equation}
where $\mu$ controls the degree of $\hat{S}_i$ . $\Tilde{Y}_t$ replaces $\hat{Y}_i$ in the subsequent summary selection process.

\section{Experiments}

\subsection{Datasets and Metrics}

We present a overview of datasets and metrics used for the two tasks in this section, respectively.

\subsubsection{Video memorability prediction related datasets and metrics.} Our experimentation involved the utilization of two datasets: Memento10k \cite{newman2020multimodal} and VideoMem \cite{cohendet2019videomem}. VideoMem encompasses videos, each accompanied by a descriptive caption elucidating the video content; however, it lacks acoustic information. This dataset supplies Short-Term (ST) and Long-Term (LT) video memorability scores. In contrast, Memento10k furnishes captions and ST scores for each video, but some videos lack acoustic content. Notably, a significant portion of videos in both datasets lacks audio, necessitating the exclusion of audio cues from our model. Figure 5 outlines the procedure for collecting ST ground-truth \cite{cohendet2019videomem, newman2020multimodal}. Specifically, human observers view a video sequence, and the identified target videos are employed to derive ST ground-truth. For instance, 100 individuals watch a designated target video and subsequently revisit it after intervals. Of the participants, 70 recall the video, while 30 do not. The ST score for this target video is computed as 70/100 = 0.7. Long-Term (LT) ground-truth is acquired over more extended periods.

To assess the model's performance in predicting video memorability, we employ the Spearman rank correlation coefficient ($RC$).

\begin{figure}[ht]
    \centering
    \includegraphics[width=\linewidth]{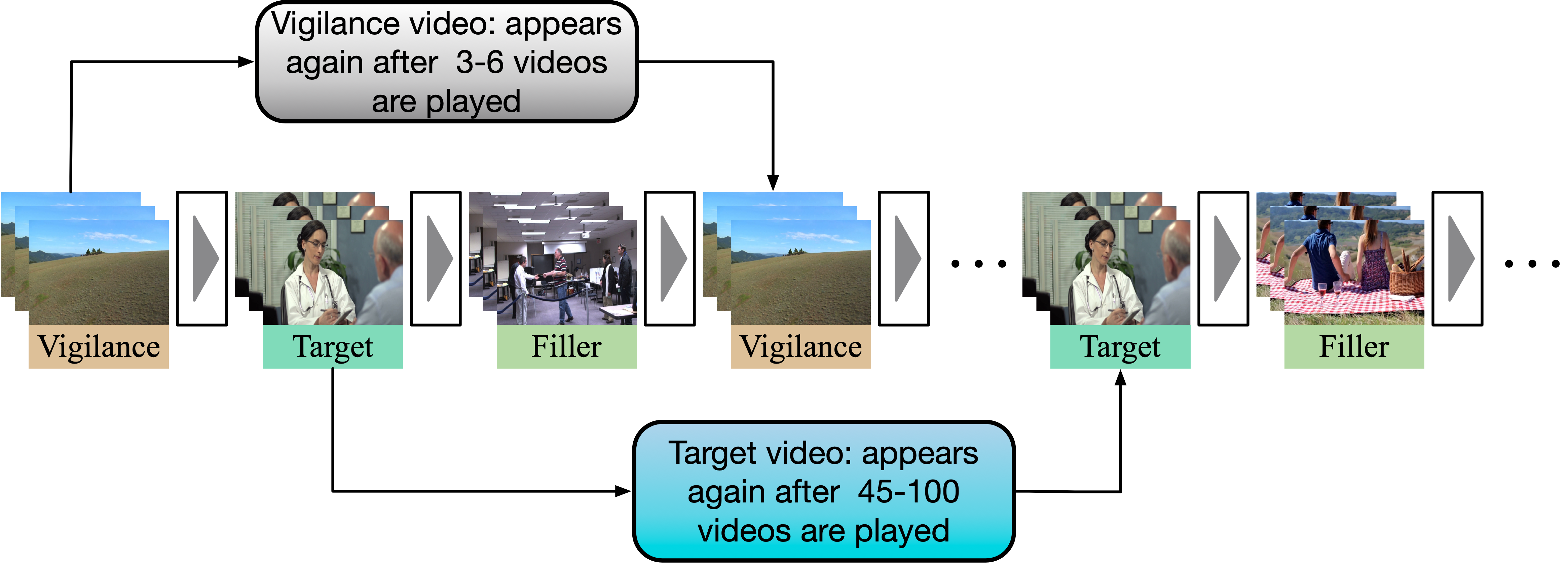}
    \caption{Protocol to collect ground-truth ST scores.}
    \label{figure5}
\end{figure}

\begin{equation}
    \rho = 1 - \frac{6\sum_{i=1}^N(\hat{S}^{(i)}-S^{(i)})}{N(N^2-1)}
    \label{equation25}
\end{equation}
where $N$ is the video number of dataset, $\hat{S}^{(i)}$ means predicted score rank of video $i$, and $S^{(i)}$ means ground-truth rank of video $i$.

\subsubsection{Video summarization related datasets and metrics.} Two benchmark datasets were involved in our experiments: SumMe \cite{gygli2014creating} and TVSum \cite{song2015tvsum}. They are labeled frame-level summary importance scores. The two datasets are tested based on 5-fold cross-validation and the reported $F_1$ scores are generated by averaging $F_1$ scores of 5 splits. $F_1$ is defined as:

\begin{equation}
    \label{equation26}
    F_1 = \frac{1}{N}\sum_{i=1}^N\frac{precision_i \times recall_i \times 2}{precision_i + recall_i}
\end{equation}
where $N$ is the long video number of dataset, $precision_i$ denotes the ratio of the overlap between the predicted video summary and the ground-truth to the predicted video summary, $recall_i$ denotes the ratio of the overlap between the predicted video summary and the ground-truth to the ground-truth.

\subsection{Implementation Details}

For video memorability prediction, we provide a comprehensive overview of the pre-trained models employed. Specifically, CLIP was "ViT-B/32", while Bert was implemented "bert-base-uncased" version. The I3D model was pre-trained on both the ImageNet and Kinetics datasets. To facilitate the training of the entire model, we adopted the Mean Squared Error (MSE) loss function, utilizing the Adam optimizer with an initial learning rate of 0.001 and weight decay set at 0.0001. The learning rate underwent adjustments through the StepLR function at equal intervals per 60 epochs. The batch size was set to 64, and the model was trained for a total of 200 epochs. The hyperparameters were configured as follows: $K$: 8, $\mathbb Q$: 1024, $\tau$: 0.07, $\lambda$: 0.5.

For video summarization, the parameters of TMCCL were pre-trained on the Memento10k dataset and subsequently frozen. The parameters of MSVA underwent training using the Adam optimizer, with an initial learning rate set at 0.0005 and weight decay at 0.0001. The value of $mu$ was designated as 0.5. The batch size was set as 4, and the model was trained for 200 epochs. Additionally, MSE was employed as the loss function for training the model.

\subsection{Experimental Results of TMCCL Multimodal Video Memorability Prediction Model}

\subsubsection{Comparison with Other Models}

Tables 1 and 2 demonstrate performance comparisons between TMCCL and alternative models across Memento10k and VideoMem datasets. Our choice of models for comparison is guided by their noteworthy performance in the aforementioned datasets, as well as their inclusion of feature extraction and fusion modules. The findings reveal that TMCCL surpasses existing models in terms of $RC$ performance, underscoring the superiority of our proposed model. Furthermore, our model achieves a new state-of-the-art for $RC$ on two datasets.

Table 1 presents an evaluation of the performance of various models on the Memento10k dataset. Notably, TMCCL demonstrates a 2\% improvement over the state-of-the-art ST-$RC$ in comparison to existing works. In Table 2, a comprehensive comparison between TMCCL and alternative models is provided, specifically focusing on the VideoMem dataset. Our model exhibits a 1\% enhancement in ST-$RC$ and a noteworthy 6.3\% improvement in LT-$RC$ when contrasted with preceding models. This performance boost is attributed to the robust visual appearance features and motion features incorporated in our model. The visual appearance features, extracted using multi-level encoding and attention methods, showcase a robust representation capability. Furthermore, the motion features, extracted through TMCCL, leverage complementary information from textual cues.

\begin{table}[ht]
  \caption{ST-$RC$ comparison on Memento10k}
  \label{table1}
  \resizebox{\linewidth}{!}{
      \begin{tabular}{cc}
        \toprule
        \hspace{1cm} Model \hspace{1cm} & \hspace{1cm} ST-$RC$ \hspace{1cm} \\
        \midrule
        \hspace{1cm} MESD \cite{newman2020multimodal} \hspace{1cm} & \hspace{1cm} 0.663 \hspace{1cm} \\
        \hspace{1cm} PMMC \cite{sweeney2020predicting} \hspace{1cm} & \hspace{1cm} 0.524 \hspace{1cm} \\
        \hspace{1cm} TOTF \cite{kleinlein2021topic} \hspace{1cm} & \hspace{1cm} 0.600 \hspace{1cm} \\
        \hspace{1cm} UVTM \cite{constantin2021using} \hspace{1cm} & \hspace{1cm} 0.648 \hspace{1cm} \\
        \hspace{1cm} FVST \cite{kleinlein2021thau} \hspace{1cm} & \hspace{1cm} 0.656 \hspace{1cm} \\
        \hspace{1cm} EMPE \cite{reboud2021exploring} \hspace{1cm} & \hspace{1cm} 0.658 \hspace{1cm} \\
        \hspace{1cm} MVME \cite{usmani2022modelling} \hspace{1cm} & \hspace{1cm} 0.661 \hspace{1cm} \\
        \hspace{1cm} PMMU \cite{constantin2022aimultimedialab} \hspace{1cm} & \hspace{1cm} 0.665 \hspace{1cm} \\
        \hspace{1cm} DSDV \cite{sweeney2022diffusing} \hspace{1cm} & \hspace{1cm} 0.667 \hspace{1cm} \\
        \hspace{1cm} M3-S \cite{dumont2023modular} \hspace{1cm} & \hspace{1cm} 0.670 \hspace{1cm} \\
        \hspace{1cm} TMCCL(\textbf{ours}) \hspace{1cm} & \hspace{1cm} \textbf{0.692} \hspace{1cm} \\
      \bottomrule
    \end{tabular}}
\end{table}

\begin{table}[ht]
  \caption{ST-$RC$ and LT-$RC$ comparison on VideoMem}
  \label{table2}
  \resizebox{\linewidth}{!}{
      \begin{tabular}{ccc}
        \toprule
        \hspace{0.5cm} Model \hspace{0.5cm} & \hspace{0.5cm} ST-$RC$ \hspace{0.5cm} & \hspace{0.5cm} LT-$RC$ \hspace{0.5cm} \\
        \midrule
        \hspace{0.5cm} SEM \cite{cohendet2019videomem} \hspace{0.5cm} & \hspace{0.5cm} 0.503 \hspace{0.5cm} & \hspace{0.5cm} 0.260 \hspace{0.5cm} \\
        \hspace{0.5cm} MDFF \cite{leyva2019multimodal} \hspace{0.5cm} & \hspace{0.5cm} 0.518 \hspace{0.5cm} & \hspace{0.5cm} 0.261 \hspace{0.5cm} \\
        \hspace{0.5cm} MESD \cite{newman2020multimodal} \hspace{0.5cm} & \hspace{0.5cm} 0.556 \hspace{0.5cm} & \hspace{0.5cm} -     \hspace{0.5cm} \\
        \hspace{0.5cm} TOTF \cite{kleinlein2021topic} \hspace{0.5cm} & \hspace{0.5cm} 0.450 \hspace{0.5cm} & \hspace{0.5cm} 0.190 \hspace{0.5cm} \\
        \hspace{0.5cm} AMEN \cite{li2022adaptive} \hspace{0.5cm} & \hspace{0.5cm} 0.604 \hspace{0.5cm} & \hspace{0.5cm} 0.259 \hspace{0.5cm} \\
        \hspace{0.5cm} M3-S \cite{newman2020multimodal} \hspace{0.5cm} & \hspace{0.5cm} 0.563 \hspace{0.5cm} & \hspace{0.5cm} -     \hspace{0.5cm} \\
        \hspace{0.5cm} TMCCL(\textbf{ours}) \hspace{0.5cm} & \hspace{0.5cm} \textbf{0.614} \hspace{0.5cm} & \hspace{0.5cm} \textbf{0.324} \hspace{0.5cm} \\
        \bottomrule
    \end{tabular}}
\end{table}

\subsubsection{Ablation Study of TMCCL}

Table 3 illustrates the impact of TMCCL on two datasets. Whether subjected to a single motion feature test or a multi-modal test, the motion features extracted by I3D with TMCCL demonstrate a notable 5\% to 10\% enhancement on both ST-$RC$ and LT-$RC$, affirming its efficacy. Inadequate training data during model fine-tuning results in a suboptimal representation of features. Text cues, serving as succinct summaries of video content, imply a semantic coherence between textual and motion cues. Consequently, these cues are employed to augment the representation of motion features. We furnish I3D with complementary information through contrastive loss. As shown in Figure 1, two videos exhibit low similarity in the motion representation space but manifest high similarity in the text representation space. Leveraging motion features in conjunction with text cues, we predict approximate video memorability scores.

\newcommand{\tabincell}[2]{\begin{tabular}{@{}#1@{}}#2\end{tabular}}

\begin{table}[ht]
    \centering
    \caption{Effect of TMCCL on Memento10k and VideoMem}
    \resizebox{\linewidth}{!}{
        \begin{tabular}{ccccc}
        \toprule
        Dataset & Feature & Method & ST-$RC$ & LT-$RC$ \\
        \midrule
        Memento10k & \tabincell{c}{ I3D-Motion  \\ I3D-Motion \\ Multi-modality \\ Multi-modality} & \tabincell{c}{  w/o TMCCL \\  w/ TMCCL \\ w/o TMCCL \\  w/ TMCCL} & \tabincell{c}{ 0.551 \\ \textbf{0.576} \\ 0.649 \\ \textbf{0.692}} & \tabincell{c}{ - \\ - \\ - \\ - } \\
        
        VideoMem & \tabincell{c}{  I3D-Motion  \\ I3D-Motion \\ Multi-modality \\ Multi-modality  } & \tabincell{c}{  w/o TMCCL \\  w/ TMCCL \\ w/o TMCCL \\  w/ TMCCL } & \tabincell{c}{ 0.394 \\ \textbf{0.417} \\ 0.542 \\  \textbf{0.614}} & \tabincell{c}{ 0.181 \\ \textbf{0.212} \\ 0.292 \\  \textbf{0.324}} \\
        \bottomrule
        \end{tabular}
        \label{table3}}
\end{table}

\subsubsection{Visulization Analysis}

Figure 6 shows the impact of TMCCL on motion features in the context of predicting video memorability scores. For our analysis, we specifically selected five videos from the Memento10k dataset and arranged them in ascending order based on their ground-truth scores, as illustrated in the initial column. The second column presents the ranking outcomes derived from motion features extracted using I3D without TMCCL, while the third column presents the results obtained with I3D incorporating TMCCL. A comparison between the second and third columns reveals that the incorporation of TMCCL enhances the representation of motion features, leading to predicted rankings that closely correspond to the ground-truth scores.

\begin{figure}[ht]
    \centering
    \includegraphics[width=\linewidth]{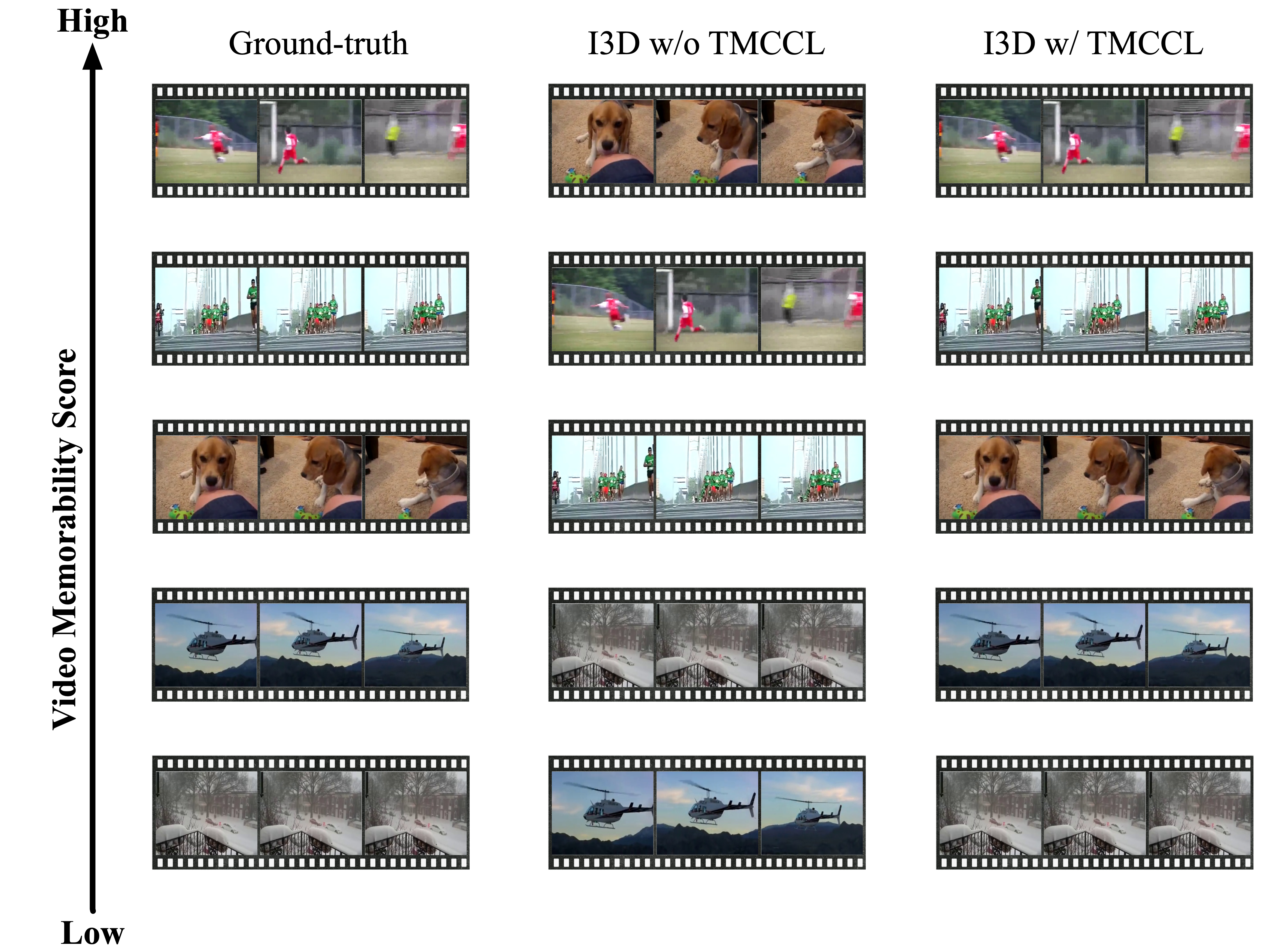}
    \caption{Qualitative analysis of the effect of TMCCL.}
    \label{figure6}
\end{figure}

\subsection{Experimental Results of Memorability Weighted Correction for Video Summarization}

\subsubsection{Comparison with Other Models} Table 4 shows the compar-
isons of MSVA with MWCVS and other models, which are dedicated to designing elaborate networks or introducing multiple categories of features. “official” means the results presented in original MSVA paper, while “our” means the results of MSVA we reproduced. We could reproduce the results on SumMe, but model performance is slightly worse on TVSum from the table. And MSVA with MWCVS improves $F_1$ performance on both datasets. There is no specific
criterion that defines interesting parts in a video. It is reasonable to introduce factors of human perception and cognition. As memora- bility is intimately related to human perception and cognition, we introduce video memorability to compensate for the weakness of existing methods, proposing MWCVS, where the rectified summary importance scores incorporate human perception and cognition factors.

\begin{table}[ht]
  \caption{Comparision with other video summarization methods on Summer and TVSum}
  \label{table4}
  \resizebox{\linewidth}{!}{
      \begin{tabular}{ccc}
      \toprule
       Model & SumMe-$F_1$(\%) & TVSum-$F_1$(\%) \\
      \midrule
      RSGN \cite{zhao2021reconstructive} & 45.0 & 60.1 \\
      DHAVS \cite{lin2022deep} & 45.6 & 60.8 \\
      3DST-UNet \cite{liu2022video} & 47.4 & 58.3 \\
      LMHA \cite{zhu2022learning} & 51.1 & 61.0 \\
      LMVS \cite{nam2023does} & 45.8 & 60.5 \\
      RUAMN \cite{su2023recurrent} & 52.3 & 60.6 \\
      MSVA(``official'') \cite{ghauri2021supervised} & 54.5 & 62.8 \\
      MSVA w/o MWCVS (ours) & 54.3 & 60.1 \\
      MSVA w/ MWCVS (ours) & \textbf{56.6} & \textbf{61.9} \\
      \bottomrule
    \end{tabular}}
\end{table}

\subsubsection{Ablation Study of MWCVS}

Table 5 shows the impact of
parameter $\mu$, which represents the weight of video memorability
scores in summary important scores. The best results are obtained
by setting $\mu$ to 0.5 for SumMe and TVSum. We observe that when
importance score on results is weakened. When the $\mu$ is too small,
$\mu$ is excessively large, the impact of the MSVA-predicted summary 860 the impact of video memorability is not enough to introduce human cognitive and perceptual factors. The balance is achieved when $\mu$ = 0.5, so we finally take $\mu$ = 0.5 in our experiments.

\begin{table}[ht]
    \centering
    \caption{The performance of MSVA with different $F_1$ value on
SumMe and TVSum}
    \resizebox{\linewidth}{!}{
        \begin{tabular}{ccc}
        \toprule
        \hspace{0.5cm} $\mu$ \hspace{0.5cm} & \hspace{0.5cm} SumMe-$F_1$(\%) \hspace{0.5cm} & \hspace{0.5cm} TVSum-$F_1$(\%) \hspace{0.5cm} \\
        \midrule
        \hspace{0.5cm} 1 \hspace{0.5cm} & \hspace{0.5cm} 55.4 \hspace{0.5cm} & \hspace{0.5cm} 61.4 \hspace{0.5cm} \\
        \hspace{0.5cm} 0.5 \hspace{0.5cm} & \hspace{0.5cm} \textbf{56.6} \hspace{0.5cm} & \hspace{0.5cm} \textbf{61.9} \hspace{0.5cm} \\
        \hspace{0.5cm} 0.1 \hspace{0.5cm} & \hspace{0.5cm} 56.3 \hspace{0.5cm} & \hspace{0.5cm} 61.5 \hspace{0.5cm} \\
        \hspace{0.5cm} 0 \hspace{0.5cm} & \hspace{0.5cm} 54.3 \hspace{0.5cm} & \hspace{0.5cm} 60.1 \hspace{0.5cm} \\
        \bottomrule
        \end{tabular}
        \label{table5}}
\end{table}

\subsubsection{Visulization Analysis}

Figure 7 shows the summary for two
videos in TVSum. Gray bars denote the ground truth, while colored
bars denote the selected clips from the long video. The selected clips
form the video summary. Pink histogram has a higher overlap with
gray histogramthan tha blue histogram and has higher $F_1$ score. It
validates the effectiveness of MWCVS.

\begin{figure}[ht]
    \centering
    \includegraphics[width=\linewidth]{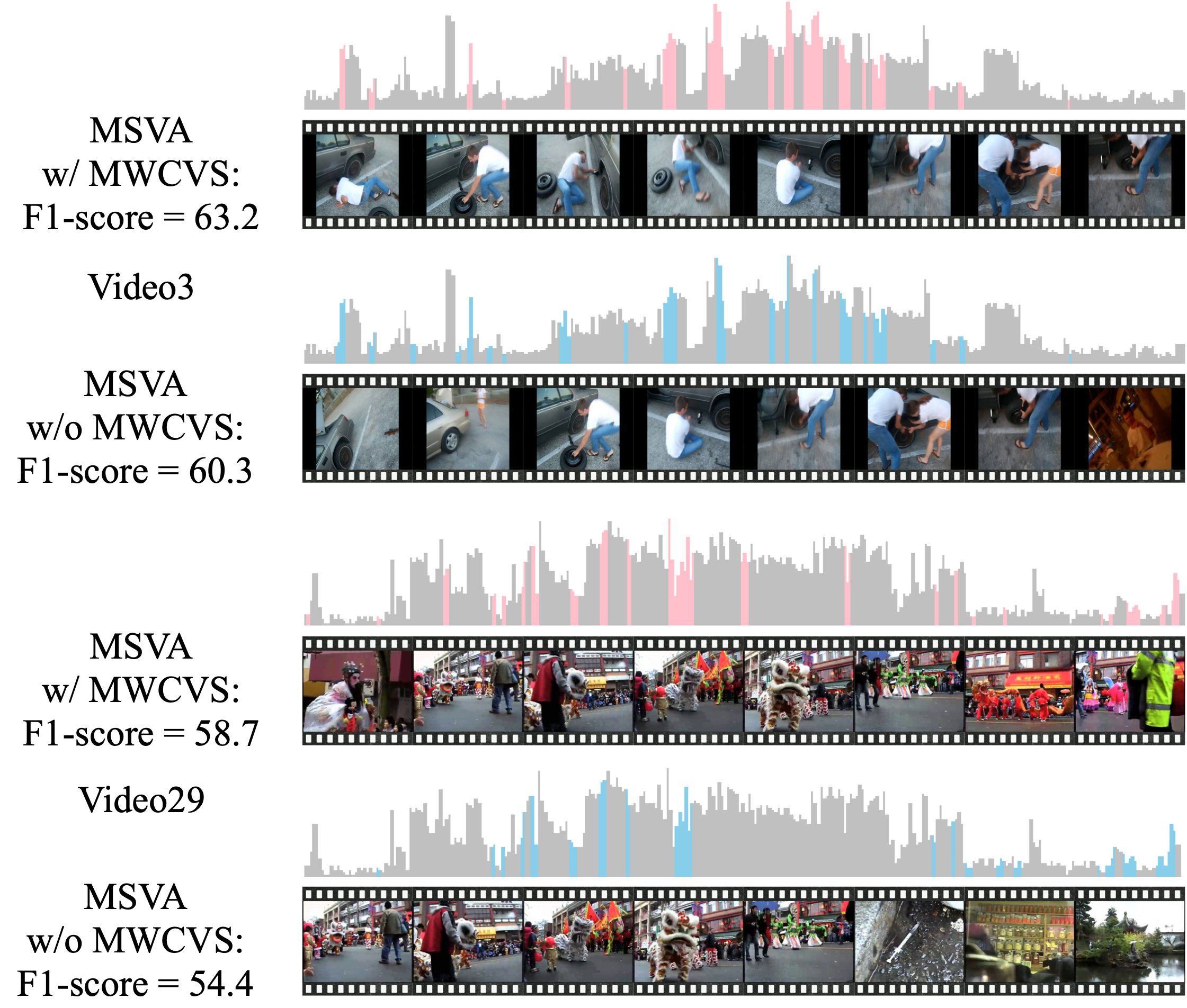}
    \caption{Qualitative comparison of MWCVS}
    \label{figure7}
\end{figure}

\section{Conclusion}

In this paper, we propose a TMCCL multimodal video memorability prediction model. Three modal features (visual appearance, text, motion) are extracted to predict video memorability scores.
However, I3D fine-tuning is limited by insufficient training data, leading to poor repretation of motion features. We propose TMCCL to provide complementary text cues for motion feature extractor fine-tuning. Our model exceeds the state-of-the-art models in $RC$ performance. Simultaneously, we propose MWCVS, introducing human cognitive and perceptual factors to address the issue of human subjectivity in video summarization labels. It achieves even better $F_1$ scores on SumMe and TVSum.

\section{Acknowledgments}
This work was supported by program in 14th Five-Year under Grant No. 2021YFF0900701,2021YFF0602103, 2021YFF0602102, 2021QY1702, and in part by Natural Science Foundation of China (No.61801441). We also thank the research funds under Grant No. 2019GQG0001 from the Institute for Guo Qiang, Tsinghua University, and the High-quality and Cutting-edge Disciplines Construction Project for Universities in Beijing (Internet Information, Communication University of China).

\section*{Declarations}

\begin{itemize}
\item Funding 
\\This work was supported by the state key development program in 14th Five-Year under Grant No. 2021YFF0900701,2021YFF0602103, 2021YFF0602102, 2021QY1702, and in part by Natural Science Foundation of China (No.61801441). We also thank the research funds under Grant No. 2019GQG0001 from the Institute for Guo Qiang, Tsinghua University, and the High-quality and Cutting-edge Disciplines Construction Project for Universities in Beijing (Internet Information, Communication University of China).

\item Conflict of interest/Competing interests (check journal-specific guidelines for which heading to use) 
\\Not Applicable

\item Ethics approval
\\Not Applicable

\item Availability of data and materials
\\Data transparency

\item Competing interests
\\The authors have no competing interests or other interests that might be perceived to influence the results and/or discussion reported in this paper.

\end{itemize}

\bibliographystyle{cas-model2-names}
\bibliography{cas-refs}

\printcredits

\end{document}